\documentclass[10pt,twocolumn,letterpaper]{article}

\usepackage{cvpr}              

\usepackage{graphicx}
\usepackage{amsmath}
\usepackage{amssymb}
\usepackage{booktabs}
\usepackage{listings}

\usepackage[pagebackref,breaklinks,colorlinks]{hyperref}

\usepackage{xcolor}
\definecolor{mylightgray}{HTML}{858585} 
\newlength\replength
\newcommand\ruleht{3pt}
\newcommand\repfrac{.3}
\newcommand\rulewidth{0.5pt}
\newcommand\drulefill{\leavevmode\dashfill\hfil%
\kern\dimexpr\repfrac\replength-\replength\relax}
\newcommand\dashfill[1][\repfrac]{\cleaders\hbox to \replength{%
\smash{\rule[\ruleht]{\repfrac\replength}{\rulewidth}}}\hfill}

\usepackage[capitalize]{cleveref}
\crefname{section}{Sec.}{Secs.}
\Crefname{section}{Section}{Sections}
\Crefname{table}{Table}{Tables}
\crefname{table}{Tab.}{Tabs.}

\def\cvprPaperID{12255} 
\def\confName{CVPR}
\def\confYear{2023}

\newcommand{\methodname}{$RoentGen$}

\begin{document}

\title{\methodname{}: Vision-Language Foundation Model for Chest X-ray Generation}

\author{Pierre Chambon\thanks{equal contribution}\\
Stanford AIMI\\
{\tt\small pchambon@stanford.edu}
\and
Christian Bluethgen\footnotemark[1]\\
Stanford AIMI\\
{\tt\small bluethgen@stanford.edu}
\AND
Jean-Benoit Delbrouck\\
Stanford AIMI\\
\and
Rogier Van der Sluijs\\
Stanford AIMI\\
\and
Ma\l{}gorzata Po\l{}acin\\
Stanford University\\
\and
Juan Manuel Zambrano Chaves\\
Stanford AIMI\\
\and
Tanishq Mathew Abraham\\
University of California, Davis\\
Stability AI\\
\and
Shivanshu Purohit\\
Stability AI\\
\and
Curtis P. Langlotz\\
Stanford AIMI\\
\and
Akshay Chaudhari\\
Stanford AIMI\\
}
\maketitle
\begin{abstract}
Multimodal models trained on large natural image-text pair datasets have exhibited astounding abilities in generating high-quality images. Medical imaging data is fundamentally different to natural images, and the language used to succinctly capture relevant details in medical data uses a different, narrow but semantically rich, domain-specific vocabulary. Not surprisingly, multi-modal models trained on natural image-text pairs do not tend to generalize well to the medical domain. Developing generative imaging models faithfully representing medical concepts while providing compositional diversity could mitigate the existing paucity of high-quality, annotated medical imaging datasets. In this work, we develop a strategy to overcome the large natural-medical distributional shift by adapting a pre-trained latent diffusion model on a corpus of publicly available chest x-rays (CXR) and their corresponding radiology (text) reports. We investigate the model's ability to generate high-fidelity, diverse synthetic CXR conditioned on text prompts. We assess the model outputs quantitatively using image quality metrics, and evaluate image quality and text-image alignment by human domain experts. We present evidence that the resulting model (\methodname{}) is able to create visually convincing, diverse synthetic CXR images, and that the output can be controlled to a new extent by using free-form text prompts including radiology-specific language. Fine-tuning this model on a fixed training set and using it as a data augmentation method, we measure a 5\% improvement of a classifier trained jointly on synthetic and real images, and a 3\% improvement when trained on a larger but purely synthetic training set. Finally, we observe that this fine-tuning distills in-domain knowledge in the text-encoder and can improve its representation capabilities of certain diseases like pneumothorax by 25\%.
\end{abstract}

\section{Introduction}
\label{sec:intro}

\begin{figure*}
  \centering
   \includegraphics[width=\linewidth]{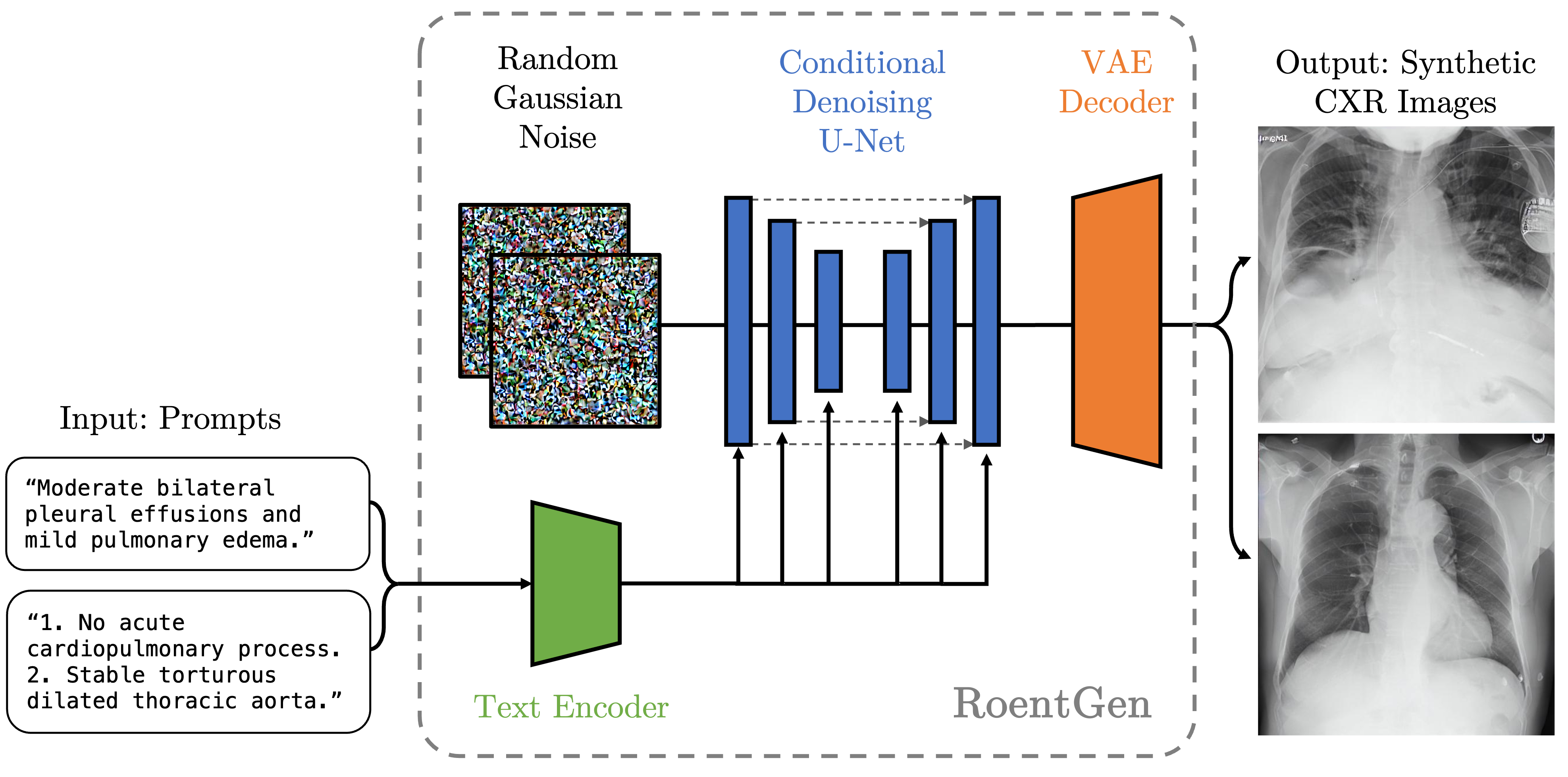}
  \caption{Text-to-image synthesis of chest x-ray images using \methodname{}, a medical domain-adapted latent diffusion model based on the Stable Diffusion pipeline. A fine-tuned or retrained conditional U-Net denoises a vector of random Gaussian noise, conditioned by embeddings created from short medical free-text prompts by a fine-tuned or replaced text encoder. The decoder of the variational autoencoder of the Stable Diffusion pipeline maps the denoised latent vector to pixel space, resulting in high-fidelity, diverse chest x-rays showing corresponding imaging features.}
  \label{fig:cherypickedexamples}
\end{figure*}

Latent diffusion models (LDMs) are a type of denoising diffusion models that have recently gained popularity by enabling high-resolution, high-fidelity and yet diverse image generation\cite{Rombach2022}. When coupled with a conditioning mechanism, these models allow fine-grained control of the image generation process at inference time (e.g., by using text prompts)\cite{ramesh2022, Rombach2022, imagen2022}. Such models have typically been trained on large, multi-modal datasets like LAION-5B which consist of billions of natural image-text pairs\cite{schuhmann2022laionb}. LDMs can be repurposed for a large variety of downstream tasks and, given appropriate pre-training, may be regarded as foundation models (FM) \cite{foundationmodels2021}. Because the denoising process of LDMs takes place in a comparatively low-dimensional latent space\cite{Rombach2022}, LDMs can be run on moderate hardware resources, facilitating their deployment to end-users. 

The impressive generative capabilities of such models permit creation of high-fidelity synthetic datasets which may be used to augment traditional supervised machine learning pipelines in scenarios that lack training data. 
This presents a possible remedy to the paucity of well-curated, annotated high-quality medical imaging datasets. Annotating such datasets requires structured planning and extensive efforts by trained medical experts capable of interpreting subtle, but semantically meaningful, image features. 

Notwithstanding the lack of large, curated, publicly available medical imaging datasets, there is typically a text-based radiology report that gives a detailed description of the relevant medical information contained in the imaging studies. This "byproduct" of medical decision making can be leveraged to automatically extract labels that can be used for downstream tasks\cite{irvin2019chexpert}, but still enforces a narrower problem formulation than could potentially be expressed with human natural language. Leveraging the vision-language entanglement of pre-trained text-conditional LDMs could provide an intuitive mechanism to create synthetic medical imaging data by prompting with relevant medical keywords or concepts of interest.

In this study, we explore the representational bounds of a large vision-language LDM (Stable Diffusion, SD) and evaluate how to adapt it to medical imaging concepts, without explicitly training on these concepts. To leverage the extensive image-text pre-training underlying the components of the SD pipeline, we explore its use for generating chest X-rays (CXR) conditioned on short in-domain text prompts. As CXRs are fast to acquire, inexpensive, and can give insight into a large variety of important medical conditions, they are one of the most widely used imaging modalities worldwide. \\

To the authors' knowledge, this is the first paper that systematically explores the domain-adaptation of an out-of-domain pretrained LDM for the language-conditioned generation of medical images beyond the few- or zero-shot setting. To this end, the representational capacity of the SD pipeline was evaluated, quantified, and ultimately expanded, exploring different strategies for improving this general-domain pretrained foundational model for representing medical concepts specific to CXRs. \\

We present \methodname{}, a generative model for synthesizing high-fidelity CXR, capable of inserting, combining and modifying imaging appearances of various CXR findings through free-form medical language text prompts, featuring highly detailed image correlates of the corresponding medical concepts. Additionally, the study establishes the following advancements: 
\begin{enumerate}

\item We present a comprehensive framework to evaluate medical domain-adapted text-to-image models using domain-specific tasks of i) classification using a pretrained classifier, ii) radiology report generation, and iii) image-image- and text-image retrieval) to assess the factual correctness of these models.

\item We compare several approaches to adapt SD to a new CXR data distribution and demonstrate that fine-tuning both the U-Net and CLIP (Contrastive Language-Image Pre-Training \cite{clip2021}) text encoder yields the the highest image fidelity and conceptual correctness.

\item The original CLIP text encoder can be replaced with a domain-specific text encoder that leads to improved performance of the resulting stable diffusion model after fine-tuning, in the setting where the text encoder is kept frozen and only the U-Net is trained.

\item The SD fine-tuning task can be used to distill in-domain knowledge to the text encoder, when trained along the U-Net, improving its representational capabilities of medical concepts such as rare abnormalities.


\item \methodname{} can be fine-tuned on a small subset (1.1-5.5k) of images and prompts for use as a data augmentation tool for downstream image classification tasks. Training with synthetic data only performed comparably to training with real data, while training jointly on real and synthetic data improved classification performance by 5\% in our setup.

\end{enumerate}

\begin{figure*}
  \centering
  \includegraphics[width=\linewidth]{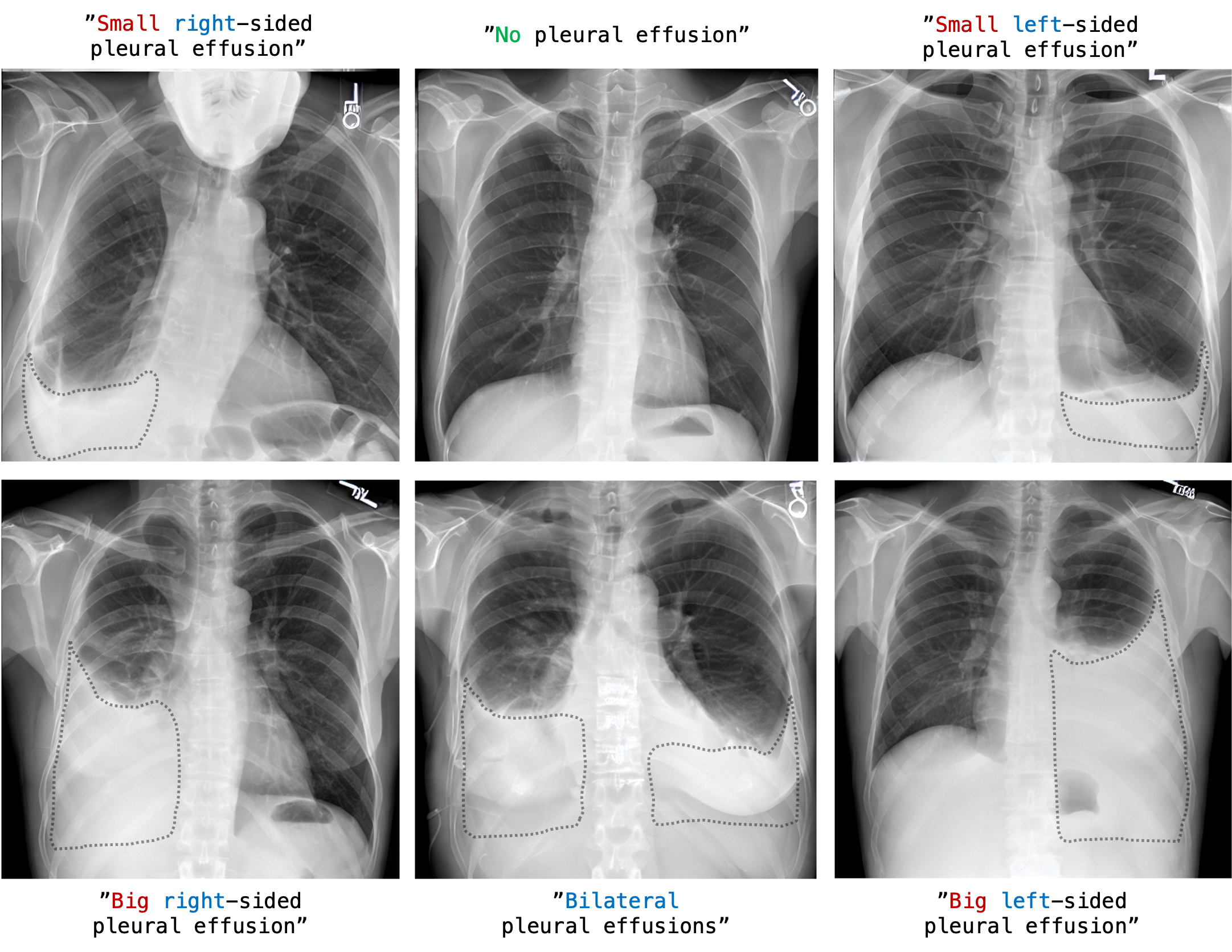}
  \caption{Text-conditioned synthesis of CXR. Each image was hand-picked out of four generated CXR per respective prompt. Here, presence or absence of a finding (pleural effusions, dotted ROI added for visualization) and dimensions like size and laterality were controlled via prompting. Note that the model correctly incorporated the radiological convention of displaying the right patient side on the left side of the image, and vice versa.}
  \label{fig:pleural_effusion_size_side}
\end{figure*}

\section{Related Work}

\subsection{Generative Models for Chest X-Ray Generation} 
Several recent works have explored the feasibility and potential benefits of synthetic CXR generation. 

A series of models have been developed to generate synthetic images for image classification of specific diseases where few training examples are available for the disease class. Such is the case of Coronavirus disease 2019 (COVID19) \cite{moris2022generation, karbhari2021generation, phukan2022covid, shams2020generative, menon2020generating, waheed2020covidgan}. \\
Other models have studied the value of synthetic CXR generation through Generative Adversarial Networks (GANs) for detecting normal (vs. abnormal) images \cite{madani2018chest}, tuberculosis  \cite{moris2022unsupervised} and pneumonia\cite{srivastav2021improved}. Other authors have studied the value of GAN-based synthetic CXRs in distinguishing multiple related chest pathologies in a multi-class classification setting. Some studies evaluated multiple closely related pathologies \cite{motamed2021data, loey2020within, shah2022dc,malygina2019data}. GAN-based synthetic CXRs have also been used for studies across multiple pathologies, identifying benefits of including synthetic CXRs in minority class performance \cite{dumont2021overcoming, sundaram2021gan}. 

Outside of image classification, other benefits of GAN-based synthetic CXR models have been described. This includes GAN-based models for inpainting \cite{ai2021pseudo, sogancioglu2018chest}, segmentation of coarse anatomical structures (lungs, heart, clavicles) \cite{ciano2021multi}, or both normal/abnormal classification and lung opacity detection \cite{tang2021disentangled}. 

Thus, most approaches have been based on generative adversarial networks (GANs), have been developed for specific pathologies and consist of single-modality (imaging only) models. However, GANs

To date, two LDMs have been described for synthetic CXR generation. The first approach demonstrated the feasibility of fine-tuning SD in a few-shot setting to generate synthetic images of single classes by text prompting, and that a CXR classifier trained on real radiographs was able to distinguish the inserted pathology with 95\% accuracy\cite{chambon2022adapting}. The second work showed the benefits of latent diffusion models in generating models across multiple individual pathologies, comparing to GAN-based approaches \cite{packhauser2022generation}. The study focused on class-conditional image generation, and compared the performance of a classifier pretrained on real CXR data for a multi-label classification task on real and synthetic data (in the latter case showing a reduced classification performance with a mean AUROC of 72\% (-9.7\%)). The authors did not report quantitative or qualitative metrics to evaluate the CXR generation.

No study was found to evaluate the benefit of LDM-based synthetic CXRs to improve downstream tasks like image classification or segmentation, and no other study attempted conditioning on text prompts. 

\subsection{LDM for other medical imaging modalities}
Two studies have used LDMs to synthesize three-dimensional imaging data for modalities like computed tomography (CT) and magnetic resonance tomography (MRI). The first study by Pinaya et al. \cite{pinaya2022ldmbrain} presented a LDM for the generation of synthetic brain MRI scans. The generative process can be conditioned on covariates like age, sex and ventricle volumes.
Another study explored the feasibility of a modified LDM for the unconditional generation CT and MRI scans \cite{khader2022medicaldiffusion}, leveraging latent space representations learned by a VQ-GAN\cite{esser2020vqgan}. The authors show that in a limited data setting, pretraining on generated data can improve the downstream performance on a segmentation task (breast MRI segmentation, Dice score 0.95 (+0.04)). However, despite training these models with a substantially lower resolutions than used in clinical practice, up to 32\% of the 50 scans evaluated by radiologists showed major anatomical inconsistencies.

\section{Datasets}

To develop a generative model capable of incorporating a variety of medical concepts formulated in natural language for the domain of CXR, we leverage the publicly available MIMIC-CXR dataset \cite{johnson2019mimic}, under institutional review board approval. The full dataset contains 377,110 images and their associated radiology reports, from 227,827 unique studies performed at the Beth Israel Deaconess Medical Center in Boston, MA, USA. \\

For the purpose of this work, this dataset was filtered using the following approach:

Each radiology report consists of two major sections: i) Findings: A textual description of normal and abnormal anatomical findings in the image, and ii) Impression: An interpretative summary of the findings for supporting medical decisions. This work focused on impression sections as input, dropping impressions shorter than 7 characters (e.g., "Slight", "Unchanged", negligible occurrence) and those exceeding the CLIP tokenizer's limit of 77 tokens (14\% of all impression sections). On the other hand, findings tend to be much longer (the last filter would drop 40\% of them) and would require text encoders with longer inputs. 

Most CXR are acquired either in a posterior-anterior (PA, detector plate in front of the patient), anterior-posterior (AP, detector plate behind the patient) or lateral (LAT) projection technique, depending on the indication, the patient's condition, and affecting the radiographic appearance of anatomical structures and pathologies. 

MIMIC-CXR contains ten approximately equally distributed and sized subgroups (p10-19). Nine subgroups (P10-P18, excluding the studies used in the official MIMIC-CXR test set) were used for training and development, while the last subgroup (p19) served as holdout set.

In each split, we cap the number of "No finding" reports, obtained through the CheXpert labeler \cite{irvin2019chexpert}, to limit the imbalance of the dataset since "No Findings" have a higher representation than all other positive findings. This enables a more balanced training and speed up the learning of positive abnormalities, especially those with rare occurrences. \\

This process yields two training sets, "PA train" (consisting exclusively of PA views) and "PA/AP/LAT" train (all views), and two test sets, "P19 test" (PA views), and MIMIC test, using the official MIMIC split. See \cref{tab:dataset} for details.

\begin{table}
  \centering
  \begin{tabular}{@{}lcccc@{}}
    \toprule
    Dataset & PA & PA/AP/LAT & MIMIC & P19 \\
    & train & train & test & test\\
    n & 38,009 & 175,622 & 2,225 & 5,000 \\
    \midrule
    \textbf{Abnormalities} & & & &\\
    ~ Atelectasis & 7795 & 40677 & 522& 918\\
    ~ Cardiomegaly & 5719 & 32744 & 520& 698 \\
    ~ Consolidation & 1779 & 9814 & 193& 182\\
    ~ Edema & 3817 & 30546 & 686& 453\\
    ~ Enlarged Card. & 1107 & 6550 & 127& 115\\
    ~ Fracture & 1420 & 4752 & 54& 142\\
    ~ Lung Lesion & 2116 & 6627 & 96& 203\\
    ~ Lung Opacity & 9136 & 41852 & 641& 1028\\
    ~ No Finding & 10005 & 47846 & 486& 1738\\
    ~ Pl. Effusion & 7859 & 41204 & 694& 1051\\
    ~ Pl. Other & 652 & 1997 & 58& 77\\
    ~ Pneumonia & 8250& 33139 & 512& 921\\
    ~ Pneumothorax & 1413 & 7304 & 62& 162\\
    ~ Sup. Devices & 3123 & 32732 & 487& 368\\
    \textbf{Impression} & & & &\\
    ~ Mean no. char. & 128.0 & 142.0 & 199.1 & 118.8\\
    ~ Std & 90.1 & 96.4 & 174.0 & 89.5\\
    ~ Mean no. tok. & 29.0 & 32.1 & 44.3 & 27.3\\
    ~ Std & 18.5 & 20.0 & 37.19 & 18.4\\
    \bottomrule
  \end{tabular}
  \caption{Composition and details of each train and test set.}
  \label{tab:dataset}
\end{table}

\label{dataset}

\section{Stable Diffusion Fine-Tuning}
\subsection{Model Architecture}


 Stable Diffusion (SD) is a powerful pipeline of several collaborative models \cite{Rombach2022}, with three main components relevant to the presented work (\cref{fig:cherypickedexamples}):
\begin{itemize}
    \item The variational autoencoder (VAE) with an encoder capable of compressing high-dimensional inputs into lower-dimensional latent representations, and a decoder mapping processed latent representations back to the pixel space.
    \item A conditional denoising U-Net, used to iteratively denoise an initially randomly generated latent vector.
    \item A conditioning mechanism. In the case of the original SD pipeline, this is a CLIP text encoder mapping text inputs to a 768-dimensional embedding space, with a limit of 77 tokens (CLIP tokenizer limit)\cite{clip2021}. 
\end{itemize}

In the following, unless otherwise specified, this architecture was not modified except for disabling the built-in "safety checker", as it was found to have a high false-positive rate for medical prompts.

\subsection{Few-shot fine-tuning}

Previous work investigated several approaches to fine-tune the SD pipeline for CXR generation in a few-shot setting\cite{chambon2022adapting}. \\  

Using a technique called "Textual Inversion", Gal et al. explored the possibility of keeping the VAE, the U-Net and almost all parts of the text encoder frozen\cite{https://doi.org/10.48550/arxiv.2208.01618}. In order to avoid catastrophic forgetting, they choose to only introduce a new token, e.g., $<$\emph{chest-xray}$>$ to describe a general image style or $<$\emph{pleural-effusion}$>$ to capture a certain object/abnormality, and train only the embedding of this token to correctly learn what it describes, on a set of images and corresponding prompts where this new token appears. \\

Another approach, called "Textual Projection" \cite{chambon2022adapting} leverages the text encoder part of the SD pipeline, to keep the existing text encoder frozen and not introduce any new token; instead replacing the original general-domain CLIP text encoder with a domain-specific text encoder followed by a projection head that maps the structure of the domain-specific embedding space to the structure of the general-domain embedding space of CLIP - used to condition the denoising process of the frozen U-Net. This approach did not produce satisfactory results in the few-shot learning setting, so that further exploration in terms of training effort and projection head is needed. \\

The "DreamBooth" approach introduced by Ruiz et al. focuses on the U-Net component, unfreezing it while keeping the VAE and text encoder frozen, and further fine-tuning on a few examples from the domain-specific images\cite{https://doi.org/10.48550/arxiv.2208.12242}. In order to avoid catastrophic forgetting and/or incrementally learn new concepts or styles, the authors suggest the use of a prior-preserving loss, where pairs of images and prompts belonging to the prior are randomly sampled from the newly generated ones at training time to maintain the performance of the model on these prior elements. Using this approach it is possible to generate synthetic, high-fidelity CXR and insert simple pathologies by text-conditioning. However, with this approach it is still easy to overfit the model and the image generation diversity is low\cite{chambon2022adapting}. The presented works use the few-shot fine-tuned "DreamBooth" model as a baseline.

\subsection{Fine-tuning and training from scratch}

In this work, we explore the potential of SD to be fine-tuned or retrained on medical domain-specific images and prompts, leveraging a large, radiology image-text dataset. In particular, for a set of images and prompts, we leverage the VAE, the text encoder and the U-Net, and compute an MSE loss that can be used to backward propagate and train the different components of the SD pipeline.

More specifically, for each text-image pair $(x_{text}, y_{pixel})$, random gaussian noise $N$ gets sampled in the latent space of dimensions $(h, w)$:

\begin{equation}
N \sim \mathcal{N}(\mathbf{0}_{h\times w}, \mathit{\mathbf{I}}_{(h\times w)^2})
\label{eq:noise}
\end{equation}

Using the text encoder and the VAE, both the prompt $x_{text}$ and the corresponding image $y_{pixel}$ are encoded, and sampled noise $N$ is added to the latent representation of the latter for a random number of timesteps $t$. The U-Net processes this noisy latent representation $\mathit{VAE}(y_{pixel})\oplus_tN$ along the encoded conditioning prompt $\mathit{Enc_{text}}(x_{text})$ to predict the original sampled noise $\hat{N}$:

\begin{equation}
\hat{N} = \mathit{Unet}(\mathit{Enc_{text}}(x_{text}), \mathit{VAE}(y_{pixel})\hspace{-0.2em}\oplus_t\hspace{-0.2em}N, t)
\label{eq:pred}
\end{equation}

An MSE loss computed between the true and predicted noises $N$ and $\hat{N}$ enables to compute gradients and improve the generation capabilities of the combined VAE, text encoder and U-Net:

\begin{equation}
\mathcal{L} = \frac{1}{h\times w}\sum_{i=0}^{h}\sum_{j=0}^{w}(\hat{N}_{i,j} - N_{i,j})^2
\label{eq:loss}
\end{equation}

Previous work indicated that the VAE component is suitable for CXR generation without modifications \cite{chambon2022adapting}. Thus, for the following experiments, the VAE was kept frozen. The experimental effort is then focused on exploring:
\begin{itemize}
    \item the U-Net component. This component is kept unfrozen, as earlier studies showed limited performance when freezing the U-Net\cite{chambon2022adapting}. It can either be further fine-tuned from the original SD work \cite{Rombach2022} (default approach), or reset randomly and trained from scratch on the in-domain dataset.
    \item the text encoder. This component can be kept frozen, therefore only training the U-Net, or it can be unfrozen and trained jointly with the U-Net (default approach). Finally, provided that the tokenizer limits are preserved according to the original SD work using a CLIP text encoder, the text encoder can be replaced with a domain-specific text encoder. In this latter case, the U-Net is trained and learns to be conditioned on this new text encoder. 
\end{itemize}

In addition to these different fine-tuning approaches, the size and distribution of the training dataset (\cref{dataset}), the number of training steps and the learning rate can vary; and finally the training can be done in full- or half-precision (fp32 and fp16) or using the Brain Floating point format (bf16). 

\subsection{Training details}

Experiments were conducted on 64 A100 GPUs split across two compute instances. Models were mostly trained (unless otherwise specified) in bf16 precision, as this led to 1/3 reduction of training time. Using bf16 precision, no significant training time difference was noticed across the various experiments, whether they fine-tuned one or several SD components, except when changing the number of training steps itself. At an image resolution of 512x512 px, an A100 GPU fine-tuning SD can hold a batch size of 8 (bf16 precision). Splitting batches across the GPUs of each compute instance, models were trained with a batch size of 256. In this setting, fine-tuning a model for 1k training steps took approximately 20 minutes; for 12.5k training steps, around 5 hours; for 60k training steps, one day.\\

Model weights for the SD pipeline (version 1.4) were obtained from the repository "CompVis/stable-diffusion-v1-4"(Hugging Face hub \cite{https://doi.org/10.48550/arxiv.1910.03771}), unless otherwise specified. The code implementation was built on both the diffusers library \cite{von-platen-etal-2022-diffusers} and the ViLMedic library \cite{delbrouck-etal-2022-vilmedic}. Two domain-specific text encoders were used: RadBERT \cite{chambon_cook_langlotz_2022}, downloadable on the HuggingFace repo \href{https://huggingface.co/StanfordAIMI/RadBERT}{StanfordAIMI/RadBERT}, and SapBERT \cite{liu-etal-2021-self} available at \href{https://huggingface.co/cambridgeltl/SapBERT-from-PubMedBERT-fulltext}{cambridgeltl/SapBERT-from-PubMedBERT-fulltext}. In the experiments that follow, guidance scale 4 and 75 inference steps with a PNDM noise scheduler \cite{https://doi.org/10.48550/arxiv.2202.09778} enabled the generation of synthetic images properly conditioned on the associated prompts.

\section{Fidelity and diversity of generated images}
\label{fidelity-diversity-images}

\begin{table}[!t]
	\centering
	\setlength{\tabcolsep}{5pt}
    \begin{tabular}{@{}lcccc@{}}
        \toprule
        &\multicolumn{3}{c}{FID $\downarrow$} & \multicolumn{1}{c}{MS-SSIM $\downarrow$}\\
        Experiments & XRV & IncepV3 & CLIP & \\
        \midrule
        \textbf{Baselines} & & & & \\
        ~ Original SD & 47.7 & 275.0 & 52.7 & .09 $\pm$ .05 \\
        ~ DreamBooth SD & 19.5 & 122.4 & 18.6 & .28 $\pm$ .07 \\
        \textbf{LR, train steps} & & & \\
        ~ 1e-4, 1k   & 6.1  & 64.6       & 2.1 & .30 $\pm$ .10 \\
        ~ 5e-5, 1k     & 6.0  & 65.0       & 2.1 & .32 $\pm$ .08\\
        ~ 1e-4, 12.5k   & 6.4  & 66.9       & 2.7 & .22 $\pm$ .10\\
        ~ 5e-5, 12.5k   & 8.2  & 67.4       & 3.3 & .19 $\pm$ .09\\
        ~ 1e-4, 60k     & 7.4  & 85.5       & 8.1 & .22 $\pm$ .09\\
        ~ 5e-5, 60k     & \textbf{3.6}  & 54.9       & 2.6 & .32 $\pm$ .09 \\
        \textbf{Components} & & & \\
        ~ Rnd U-Net, 1k             & 9.0  & 233.8      & 17.8 & .20 $\pm$ .08\\
        ~ Rnd U-Net, 60k            & 4.9  & 75.4       & 2.8 & .29 $\pm$ .09 \\
        ~ Rnd U-Net only,  & & &\\
        ~ 60k & 16.5 & 114.2      & 7.0 & .14 $\pm$ .06 \\
        ~ U-Net only, 60k        & 9.2  & 85.5       & 4.2 & .19 $\pm$ .09\\
        \textbf{Text Encoders} & & & & \\
        ~ RadBERT, 1k                      & 8.3  & 227.2      & 19.1 & .21 $\pm$ .10\\
        ~ RadBERT, 12.5k                   & 4.6  & 68.5       & 6.0 & .26 $\pm$ .12  \\
        ~ SapBERT, 60k       & 6.0  & 72.0       & 3.0 & .22 $\pm$ .10\\
        ~ RadBERT, 60k       & 6.7  & 88.3       & 5.9  & .19 $\pm$ .09\\
        \textbf{Multiple Views} & & & \\
        ~ 60k             & 19.3 & 114.0      & 5.4 & .12 $\pm$ .07\\
        \bottomrule
    \end{tabular}
\caption{Quantitative assessment of image fidelity and diversity: FID for features extracted from a CXR-pretrained DenseNet-121 (XRV), ImageNet-pretrained InceptionV3, and natural image-text-pretrained clip-ViT-B-32 (CLIP). Smaller FID indicates higher fidelity to the original images. Smaller MS-SSIM indicates higher intra-prompt image generation diversity. SD: Stable Diffusion. Image diversity: MS-SSIM mean and standard deviation. Fréchet Inception Distance (FID). MS-SSIM: Multi-scale structural similarity index measure.}
\label{tab:fid_results}
\end{table}

Generative models need to show two major qualities: the generated images should be close to the distribution they are modeled after (fidelity), and the outputs should ideally cover a large variability of the underlying real images (diversity). To assess fidelity, the Fréchet Inception Distance (FID) was calculated as a metric for how similar the distributions of real and synthetic images are. A smaller FID indicates that generated images are more similar to original images, and vice versa.
As FID is typically calculated using an Inception V3 model\cite{szegedy2015} pretrained on ImageNet, it might fail in a domain adaptation setting by being unable to capture relevant features of the CXR modality\cite{kynkaanniemi2022fid}. For this reason, FID scores were calculated from intermediate layers of three fundamentally different (in terms of architecture, domain, and pretraining) models: InceptionV3 (2048-dimensional activation vector from the pool3 layer) pretrained on ImageNet, CLIP-ViT-B-32 (768-dimensional) trained on millions of image-text pairs\cite{clip2021}, and an in-domain classification model trained to detect common pathologies in CXR (DenseNet-121, XRV, 1024-dimensional)\cite{Cohen2022xrv}.

Generation diversity was assessed by calculating the pairwise multi-scale structural similarity index metric (MS-SSIM, Gaussian kernel size 11; sigma, 1.5)\cite{wang2003multiscale} of four generated samples per prompt. A lower MS-SSIM indicates a smaller structural similarity between images and can be interpreted as higher diversity among the outputs.

The calculated metrics allow a comparison of the quality (in terms of fidelity and diversity) of synthetic images generated from the impression sections of the p19 test set with the original images from the same test set, as displayed in \cref{tab:fid_results}. This works explores three main experimental dimensions: the effect of hyper-parameters like learning rate and training steps; which SD component best to fine-tune or retrain individually or in combination with other components; and switching the CLIP text encoder with different domain-specific text encoders.

\paragraph{Learning rate, train steps} 1k training steps improved the FID scores on the two baseline approaches, original SD and DreamBooth SD, underlining the ability of the SD pipeline to quickly learn domain specific content. As the number of training steps grew to 12.5k, FID scores slightly deteriorated, which we hypothesize to be due to the model not overfitting a few images anymore and diversifying its synthetic generations, but losing in quality for each of these generations. Finally, 60k steps provided the best quality of results when using the learning rate $\text{5e-5}$, with an FID\textsubscript{XRV} of 3.6 and an FID\textsubscript{IncepV3} of 54.9. The results of the same model evaluated with CLIP were not coherent: the FID score was higher than training with only 1k steps, though the domain-specific FID\textsubscript{XRV} coupled with a manual review strongly suggested an improvement of performance. We believe this shows the limitations of using a CLIP-based metric for this domain-specific task, once models become good enough and spotting differences and improvements need a more fine-grained evaluation. The gap in performance between 12.5k and 60k steps suggests that more training steps could further improve the quality of the synthetic distribution of images. As results were worse for a learning rate 1e-4, other experiments relied on a learning rate of 5e-5.

\paragraph{Components} All experiments on learning rates and training steps relied on jointly and continuously fine-tuning both the U-Net and the text encoder, starting from the original SD weights. This approach is compared with only fine-tune the U-Net and/or initialize the U-Net to train from scratch. Over 1k training steps, randomly initializing the U-Net and training it along the text encoder led to a 50\% deterioration compared to the continuous fine-tuning equivalent, but was still able to beat the baselines by a large margin, from FID\textsubscript{XRV} values of 47.7 and 19.5 to an FID\textsubscript{XRV} of 9.0. After 60k steps, the randomly-initialized U-Net variant achieves an FID\textsubscript{XRV} of 4.9, which does not beat an FID\textsubscript{XRV} of 3.6 achieved by our best model. Over the same number of training steps, training the U-Net only, from a random initialization, showed limitations and only achieved an FID\textsubscript{XRV} of 16.5, whereas training the U-Net only from the original SD approach yielded an FID\textsubscript{XRV} of 9.2. This underlines the importance of fine-tuning the text encoder along the U-Net, to not only speed up the learning process but also to achieve better asymptotic performance. 

\paragraph{Text Encoders} In our last main set of experiments, we explored how switching the CLIP text encoder with a domain-specific text encoder, such as RadBERT \cite{chambon_cook_langlotz_2022} or SapBERT \cite{liu-etal-2021-self}, could improve performance, in the setting where we keep the domain-specific text encoder frozen and only train the U-Net initialized from scratch. After 60k training steps, the model using SapBERT achieved an FID\textsubscript{XRV} of 6.0, with RadBERT an FID\textsubscript{XRV} 6.7, compared to the random U-Net only model that only scored an FID\textsubscript{XRV} of 16.5. This 60\% improvement in FID\textsubscript{XRV} strongly suggests that domain-specific knowledge already encoded in these text encoder can be leveraged by the SD model to generate accurate images. As a next step experiment, we would like to explore how fine-tuning these domain-specific text encoders along the U-Net could help further improve performance, while limiting catastrophic forgetting phenomenons. 

\paragraph{Diversity} All above-mentioned models were assessed using the MS-SSIM metric, with a smaller values representing a lower structural similarity, which can be interpreted as an indicator of higher diversity, and vice versa. This metric, as is, can show limitations: it indicates higher diversity for samples generated by a model trained for 1k steps (lr, 5e-5) than the same model trained for 60k steps. Manual inspection indicates that the 60k-steps variant features a richer generation diversity, but a model trained for only for 1k steps produces frequent errors and outputs out-of-domain images, achieving poor fidelity but higher MS-SSIM score. Therefore, MS-SSIM scores and FID values need to be considered jointly. 
In these experiments, we included a model trained on multiple views (PA, AP and LATERAL) for 60k steps. Although it achieved lower scores than most models in \cref{tab:fid_results}, the evaluation setting involves the p19 test set that only includes PA images, therefore partially invalidating the use of FID scores for this model. As expected, the multi-view model showed the lowest MS-SSIM among all the fine-tuned models, as it generates images from all available views. \\

\begin{figure}
  \centering
   \includegraphics[width=\linewidth]{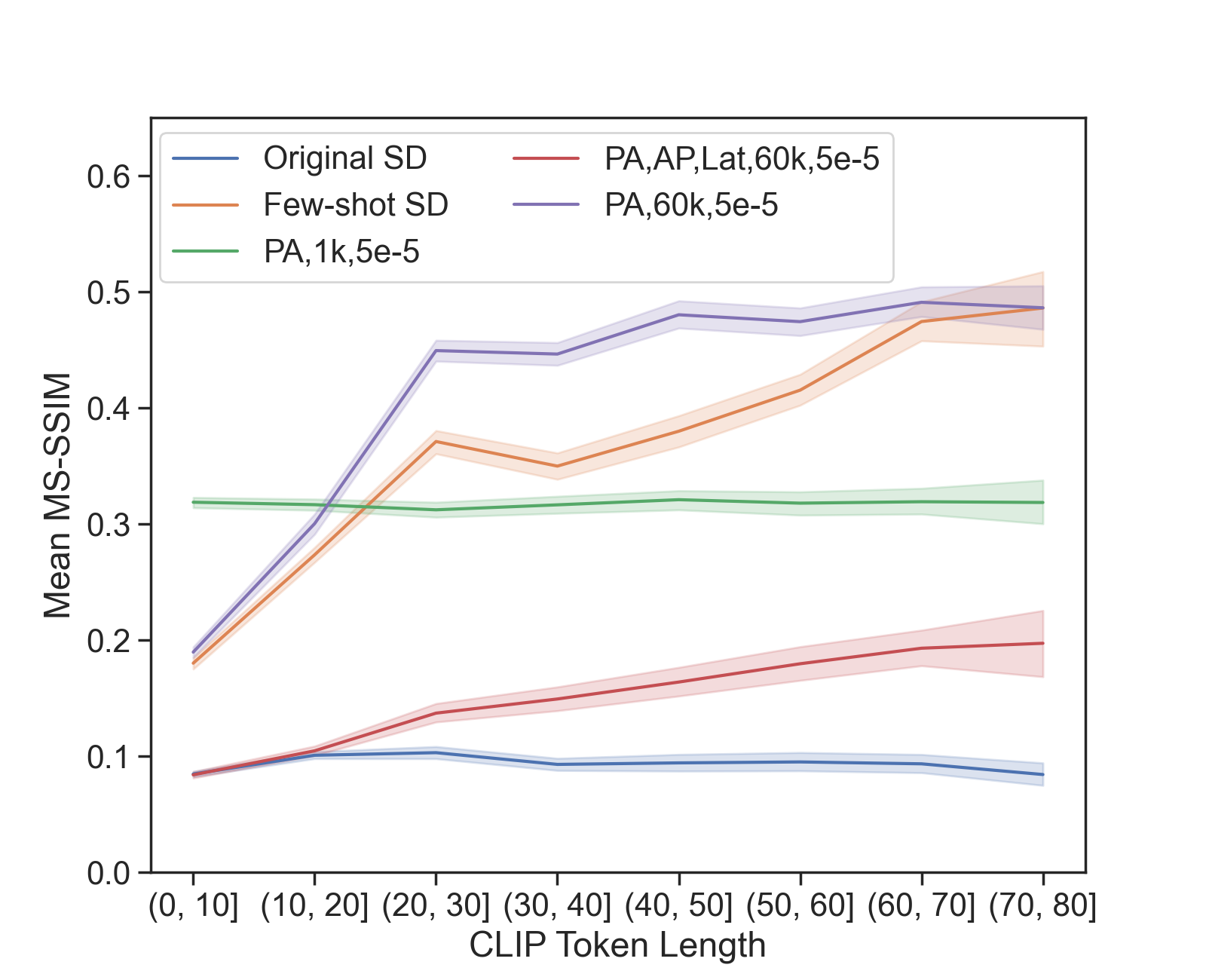}
  \caption{Intra-prompt image diversity by CLIP token length. Mean MS-SSIM as a measure of intra-prompt generation diversity for 5,000 prompts (with 4 generated images per prompt) for selected models. Lower mean MS-SSIM indicates higher diversity. For visualization, CLIP token lengths have been binned to intervals of size 10. The light areas indicate 95\% confidence intervals. SD: Stable Diffusion. PA: Postero-anterior view. AP: Antero-posterior view. Lat: Lateral view.}
  \label{fig:msssim-token-diversity}
\end{figure}

To evaluate a possible relationship between generation diversity and the length of the conditioning text prompt, we calculated the diversity (as expressed by MS-SSIM) for pairwise comparisons between generated CXR images of the same prompt \cref{fig:msssim-token-diversity}. Compared to the original SD model with a consistently high diversity (low MS-SSIM) across CLIP token lengths, the fine-tuned models tend to exhibit a smaller diversity with increasing token length. We hypothesize this to be due to higher constraints imposed by a more detailed and thus more specific prompt. This relationship is not observed for the model trained for only 1k steps, which could be explained by the above-mentioned limitation of MS-SSIM as an indicator of diversity.

\begin{figure*}
  \centering
   \includegraphics[width=\linewidth]{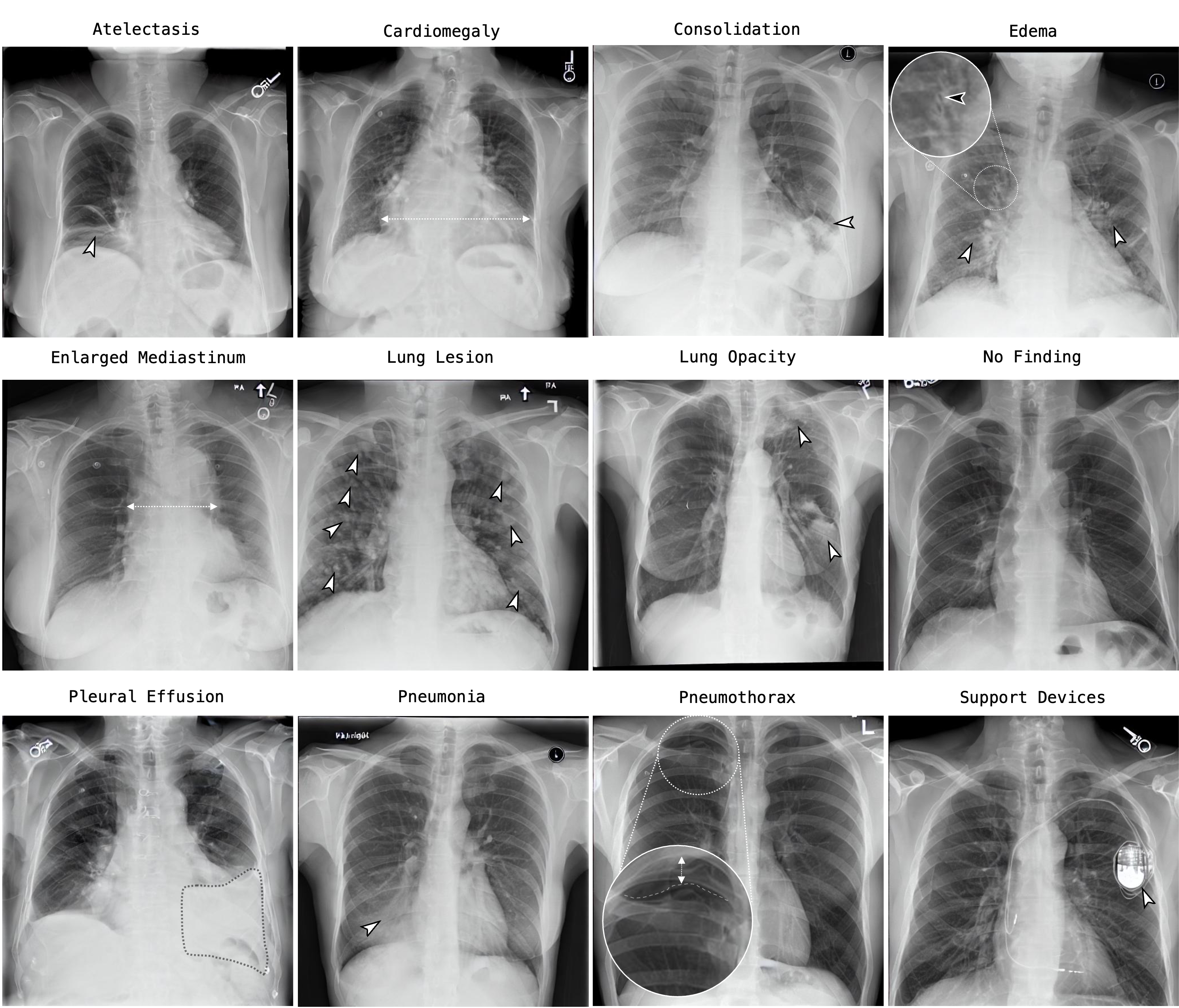}
  \caption{Synthetic images created by prompting a fine-tuned model (60k training steps; learning rate 5e-5; PA-view) for typical CXR abnormalities. The generated CXRs feature high levels of detail: When prompted for "edema" (top right), perihilar haziness(white arrowheads) and peribronchial cuffing (black arrowhead), both features seen in pulmonary edema, can be observed. For 'pneumothorax' (bottom row, third image from the left), a fine line representing the visceral pleural lining of the partially collapsed lung can be delineated (dashed line).}
  \label{fig:14lungdiseases}
\end{figure*}

\begin{table*}[]
\begin{tabular}{@{}lccccccccccccc@{}}
\toprule
Source    & Atel. & Cmgl. & Cnsl. & Edma & Les. & Opac. & Eff. & Other & Pneum. & PTX & Filtered Avg \\ \midrule
\textbf{Baselines} & & & & & & & & & & & \\
~ p19 Test Set  & .75  & .84 & .70   & .84  & .65  & .68   & .87 & .70 & .60  & .78 & .67  \\
~ Original SD & .48	&.49	&.47	&.49	&.56	&.51	&.48	&.56	&.51	&.55	&.50  \\
~ DreamBooth SD  &.62	&.54	&.58	&.50 &.66	&.56	&.72	&.70	&.51	&.66	&.61  \\
\textbf{LR, train steps} & & & & & & & & & & & \\
~ 1e-4, 1k   &.76	&.82	&\textbf{.69}	&\underline{\textbf{.85}}	&\textbf{.60}	&\underline{\textbf{.74}}	&\textbf{.90}	&.63	&\underline{\textbf{.61}}	&\underline{\textbf{.84}}	&\underline{\textbf{.84}} \\
~ 5e-5, 1k  &.75	&.79	&.66	&\textbf{.84}	&.57	&.68	&\underline{\textbf{.91}}	&.61	&.57	&\textbf{.80}	&\textbf{.82}  \\
~ 1e-4, 12.5k  &\textbf{.77}	&\textbf{.83}	&\underline{\textbf{.70}}	&.82	&\underline{\textbf{.62}}	&\textbf{.72}	&\underline{\textbf{.91}}	&\underline{\textbf{.76}}	&.56	&.79	&\textbf{.82}  \\
~ 5e-5, 12.5k  &.72	&.77	&\underline{\textbf{.70}}	&.77	&.57	&.69	&.88	&.72	&\textbf{.58}	&.76	&.78  \\
~ 1e-4, 60k &.72	&.82	&.61	&.69	&.51	&.57	&.86	&\textbf{.75}	&.45	&.73	&.77  \\
~ 5e-5, 60k  &\underline{\textbf{.78}}	&\underline{\textbf{.84}}	&.66	&.78	&.54	&.66	&.89	&.71	&.47	&.76	&.81  \\
\textbf{Components} & & & & & & & & & & & \\
~ Rnd U-Net, 1k  &.72	&.69	&.66	&.70	&\underline{\textbf{.63}}	&\textbf{.68}	&.82	&.59	&\textbf{.52}	&\textbf{.76}	&.74  \\
~ Rnd U-Net, 60k  &\underline{\textbf{.79}}	&\underline{\textbf{.82}}	&\underline{\textbf{.70}}	&\underline{\textbf{.78}}	&\underline{\textbf{.63}}	&\underline{\textbf{.73}}	&\underline{\textbf{.90}}	&\textbf{.73}	&\textbf{.52}	&\underline{\textbf{.79}}	&\underline{\textbf{.82}}  \\
~ Rnd U-Net only, 60k &.72	&\textbf{.76}	&\underline{\textbf{.70}}	&.72	&\textbf{.60}	&.67	&\textbf{.87}	&.71	&\underline{\textbf{.58}}	&\textbf{.76}	&.77 \\
~ U-Net only &\textbf{.77}	&\underline{\textbf{.82}}	&\textbf{.67}	&\textbf{.77}	&.55	&.66	&\underline{\textbf{.90}}	&\underline{\textbf{.74}}	&.48	&.74	&\textbf{.80}  \\
\textbf{Text Encoder} & & & & & & & & & & & \\
~ RadBERT, 1k &.67	&.59	&\textbf{.65}	&.65	&\textbf{.63}	&.59	&.70	&.43	&.51	&\textbf{.78}	&.68 \\
~ RadBERT, 12.5k  &\textbf{.77}	&\textbf{.78}	&.\textbf{65}	&\underline{\textbf{.80}}	&\underline{\textbf{.66}}	&\textbf{.69}	&\textbf{.87}	&\underline{\textbf{.72}}	&\underline{\textbf{.61}}	&\underline{\textbf{.81}}	&\underline{\textbf{.81}}  \\
~ RadBERT, 60k  &.72	&.73	&.63	&.67	&.62	&.68	&.86	&.66	&\textbf{.55}	&\textbf{.78}	&\textbf{.75}\\
~ SapBERT, 60k  &\underline{\textbf{.80}}	&\underline{\textbf{.84}}	&\underline{\textbf{.68}}	&\textbf{.77}	&\textbf{.63}	&\underline{\textbf{.72}}	&\underline{\textbf{.90}}	&\textbf{.71}	&.48	&.77	&\underline{\textbf{.81}} \\
\textbf{Multiple Views} & & & & & & & & & & & \\
~ 60k  &.72	 &.73	&.63	&.67	&.62	&.68	&.86	&.66	&.55	&.78	&.75  \\ \bottomrule
\end{tabular}
\caption{Classification performance of DenseNet121 on the p19 test set and on 5000 synthetic CXR generated using the models listed in the experiment column, measured by the area under the ROC curve. Per model, the impression sections corresponding to the 5000 images in the p19 test set were used to generate the sample images. Atel.: Atelectasis, Cmgl.: Cardiomegaly, Cnsl.: Consolidation, Edma: Edema, Les.: Lung lesion, Opac.: Lung Opacity, Eff.: Pleural effusion, Pneum.: Pneumonia, PTX: Pneumothorax. The scores corresponding to Enlarged Cardiomediastinum and Fracture were not reported, as the baseline p19 test set had AUROC values below or equal to 0.5. The filtered average scores is a macro-average of classes with baseline p19 test set AUROC values above 0.75, namely Atelectasis, Cardiomegaly, Edema, Pleural Effusion and Pneumothorax. In each section, class-wise best values are underlined and emboldened, and second best values are emboldened only.}
\label{tab:classification_synth}
\end{table*}

\section{Factual correctness of generated images}
While FID gives an insight of the feature discrepancy between ground-truth and generated images, it also has some important shortcomings. Notably, models  such as Inception V3 or XRV DenseNet are vision-only and trained on a classification tasks. They may not capture every fine-grained feature required to evaluate the semantic correctness of the generated images. To further test our generative models, we leveraged pre-trained multimodal models to benefit from an evaluation at the intersection of vision and language. More precisely, we used models that either generate text from images or encode medical text to report semantic, fine-grained evaluations.

\begin{table*}[!t]
	\centering
	\setlength{\tabcolsep}{3.5pt}
	\begin{tabular}{lccccccc}
		\multicolumn{1}{c}{\bf Task and models}  &\multicolumn{6}{c}{\bf 				Factual correctness metrics}
		\\ \hline \\
		&BL4$\uparrow$ & ROUGEL$\uparrow$ & $\text{F}_1$cXb$\uparrow$ &BERTScore$\uparrow$  &  $\text{fact}_\text{ENT}$$\uparrow$ & $\text{fact}_\text{ENTNLI}$$\uparrow$ &  RadGraph$\uparrow$ \\
		\textbf{Radiology Report Generation} \\
 		Ground truth images & 8.9 & 23.0 & 50.4 & 45.6 & 27.6 & 23.3 & 22.5   \\
		\multicolumn{8}{c}{{\color{mylightgray}\drulefill \drulefill}} \\
		DreamBooth SD & 2.4 & 12.8 & 36.6 & 36.6 & 13.7 & 6.9 & 9.0  \\
		5e-5, 1k & 3.6 & 15.8 & 46.2 & 38.0 & 22.9 & 14.3 & 15.0  \\
		5e-5, 60k & \textbf{5.5} & 19.9 & \underline{\textbf{46.5}} & \textbf{42.3} & \textbf{24.1} & 19.9 & 18.9  \\
		RadBERT, 60k & 4.7 & 18.0 & 42.6 & 40.3 & 20.6 & 17.4 & 16.3  \\
		SapBERT, 60k & 4.9 & \textbf{20.3} & 46.2 & \textbf{42.3} & 23.9 & \textbf{20.2} & \textbf{19.0} \\
		U-Net only, 60k & 4.1 & 15.6 & 36.3 & 38.7 & 17.4 & 13.8 & 13.5  \\
		Multiple views, 60k & \underline{\textbf{7.0}} & \underline{\textbf{22.5}} & \textbf{46.3} & \underline{\textbf{43.2}} & \underline{\textbf{24.5}} & \underline{\textbf{21.5}} & \underline{\textbf{20.1}}  \\
		\hline \\
		& Prec$@5$$\uparrow$ & Prec$@10$$\uparrow$ & Prec$@50$$\uparrow$ \\
		\textbf{Image-Image Retrieval} \\
      	Ground truth images & 55.2 & 47.8 & 40.1   \\
		\multicolumn{8}{c}{{\color{mylightgray}\drulefill \drulefill}} \\
		DreamBooth SD & 27.8 & 23.9 & 20.8 \\
		5e-5, 1k &  39.6 & 35.5 & 26.9 \\
		5e-5, 60k & \textbf{46.1} & \textbf{44.4} & \textbf{34.9}  \\
		RadBERT, 60k & 44.2 & 36.3 & 29.9  \\
		SapBERT, 60k & 42.3 & 38.8 & 32.2  \\
		U-Net only, 60k & 40.3 & 35.5 & 28.3  \\
		Multiple views, 60k & \underline{\textbf{49.5}} & \underline{\textbf{47.7}} & \underline{\textbf{39.5}}  \\
				\hline \\
		& Prec$@5$$\uparrow$ & Prec$@10$$\uparrow$ & Prec$@50$$\uparrow$ & $\text{F}_1$cXb$\uparrow$ & BERTScore$\uparrow$ &  $\text{fact}_\text{ENT}$$\uparrow$ & RadGraph$\uparrow$\\
		\textbf{Image-Text Retrieval} \\
		Ground truth images & 41.2 & 40.9 & 37.6 & 49.1 & 35.8  & 19.2 & 11.3   \\
		\multicolumn{8}{c}{{\color{mylightgray}\drulefill \drulefill}} \\
		DreamBooth SD & 22.1 & 20.4 & 17.1 & 35.0 & 31.8 & 11.4 & 6.4  \\
		5e-5, 1k & 29.3 & 28.1 & 27.1 & 42.7 & 33.8 & 13.0 & 6.8   \\
		5e-5, 60k & 35.0 & \textbf{34.4} & \textbf{30.2} & 45.3 & \textbf{34.3} & \textbf{15.3} & \textbf{9.2}\\
		RadBERT, 60k & 29.3 & 28.1 & 25.6 & 41.6 & 33.2 & 13.0 & 7.9    \\
		SapBERT, 60k & \textbf{36.5} & 32.7 & 30.1 & \textbf{45.4} & 34.0 & 14.0 & 8.2    \\
		U-Net only, 60k & 28.3 & 26.5 & 23.7 & 39.0 & 32.6 & 13.1 & 8.0    \\
		Multiple views, 60k & \underline{\textbf{40.8}} & \underline{\textbf{38.8}} & \underline{\textbf{35.2}} & \underline{\textbf{47.4}} & \underline{\textbf{34.4}} & \underline{\textbf{17.9}} & \underline{\textbf{10.2}}    \\
		\end{tabular}      
	\caption{Factual correctness metrics for the tasks of radiology report generation, image-image retrieval and image-text retrieval. Best score for each metric and each task are emboldened and underlined, second best score only emboldened.}
    \label{score-tabular-fact-orient}
\end{table*}%

\subsection{Multi-label classification}

MIMIC-CXR provides class labels for findings commonly encountered in CXR, derived by natural language processing, using the corresponding text reports similar to the work by Irivin et al.\cite{irvin2019chexpert}. Using the impression sections from the p19 test set (filtered by the above-mentioned criteria), different fine-tuned SD models were queried to produce synthetic images that reflect the abnormalities of the corresponding impression sections, as labeled by CheXpert. Such synthetic images are displayed in \cref{fig:14lungdiseases}. Without being further modification, a pre-trained classification model (DenseNet-121, XRV) was used to classify both the real images (the p19 test set baseline) as well as 5,000 images (one image per prompt) per fine-tuned SD model. NLP-derived labels of the respective reports served as ground truth. All images were preprocessed by downsizing to a maximum height or width of 512 px (preserving aspect ratio), center cropping (512 px) and downsizing to a resolution of 224 x 224 px (following the original model's specifications).\\

Putting all of this together, \cref{tab:classification_synth} displays the results after having the DenseNet-121 classification model predict classes from images generated by various SD models with the original CheXpert weak labels. As a baseline check, the original SD pipeline yields an AUROC of approximately 0.5, as the generated images are completely out of distribution. Similarly, a few-shot trained model ("DreamBooth SD") only scores a (filtered average) AUROC of 0.61. Although DreamBooth allows rapid and compute-efficient fine-tuning, this approach does yield images with accurate medical content outside of specific two-class-conditional training.

Different fine-tuned SD models can be compared using these scores, though they lead to partially incoherent conclusions. Looking at the experiments on the learning rate and the training steps, fine-tuning with $lr=\text{5e-5}$ for 1k steps achieved a filtered average AUROC of 0.82  versus an AUROC of 0.81 for 60k steps, the difference being even larger with  $lr=\text{1e-4}$. As a manual review suggests, training longer drastically increased performance. We identified several hypothesis that can explain these results:
\begin{itemize}
    \item Models trained for 1k steps systematically output images where abnormalities are represented in an obvious manner (i.e. overfit), whereas models trained longer learn more subtle representations that are harder to classify. 
    \item The CheXpert labels, used as the ground-truth, are noisy \cite{https://doi.org/10.48550/arxiv.2004.09167} and the confidence intervals exceed in width the differences between the various scores observed in \cref{tab:classification_synth}.
    \item The CheXpert labels, in the way they were generated, can be incorrect. A model trained for a short amount of time on impression sections (used to generate CheXpert labels) can make the same mistakes as the CheXpert labeler in understanding the presence and absence of abnormalities, that a model trained longer will more accurately capture.
    \item The pretrained Densenet-121 model has limited performance on the test data, as underlined by the baseline p19 test set. In particular, its macro-averaged performance on this set for the categories present in \cref{tab:classification_synth} is only 0.74. 
\end{itemize}

Taking these elements into account, we assess that such metrics would benefit from better ground-truth CheXpert labels \cite{https://doi.org/10.48550/arxiv.2004.09167}, and also warrant further exploration of other classifiers. Despite this, the findings are sufficient to capture differences between in- and out-of-domain models. The following subsections investigate other downstream applications that can be used to assess the medical correctness of synthetic images.

\subsection{Radiology Report Generation}\label{sec:fact_orient_rrg}

The task of Radiology Report Generation (RRG) consists of building assistive systems that take X-ray images of a patient and generate a textual report (called impression) describing clinical observations in the images. In the scope of our evaluation of the factual correctness of generated images, we leveraged a model pre-trained on MIMIC-CXR, which contains image-impression pairs\cite{johnson2019mimic}. Our evaluation was carried out as follows: 1) we generated images using our models for every ground-truth impression of the MIMIC-CXR test set 2) we input these generated images in the pre-trained model that outputs new impressions 3) we compare these new impressions with the ground-truth impression used to generate the images. We also report the RRG model's performances when using the ground-truth images as an upper-bound baseline. \\ 

To evaluate the new impression against the ground-truth impressions, we report the widely used natural language generation metrics such as BLEU~\cite{papineni2002bleu}, ROUGE~\cite{lin2004rouge} and BERTScore~\cite{zhang2019bertscore}. We also leverage the newly introduced factual-oriented metrics, namely the {$\text{fact}_\text{ENT}$}, $\text{fact}_\text{ENTNLI}$~\cite{miura2021improving} and $\text{RadGraph}$ score~\cite{delbrouck2022improving}. These three metrics evaluate the factual correctness, completeness and consistency of the generated impression using Named Entity Recognition and Entity Relation systems trained on radiologist-annotated data (more information in Appendix~\ref{app:detail_fact_oriented}). Finally, we report the {F$_1$CheXbert}~\cite{zhang2020optimizing} metric computing the F1-score between the prediction of CheXbert~\cite{smit2020combining} run on the generated report and the ground truth report.\\

The results in \cref{score-tabular-fact-orient} underline the capabilities of such metrics to capture the differences in generation capabilities of our models, when a fixed RRG model is used on top of the synthetic images to generate synthetic reports to be compared to the ground-truth reports. As a sanity check, we can observe the progress made from the DreamBooth SD baseline, achieving 13.7 {$\text{fact}_\text{ENT}$} and 9.0 $\text{RadGraph}$, to our model trained for 1k steps, improving these scores to 22.9 and 15.0, and finally the scores of the 60k steps variant, respectively 24.1 and 18.9. Using these RRG metrics, we notice the superiority of the approach trained on multiple views for 60k train steps: a contrario to the evaluation framework of \cref{fidelity-diversity-images}, the RRG metrics are image-view agnostic, and therefore capable of measuring the added-value of training on a more diverse set of images from multiple views. Our variants trained only on PA images, either by continuously fine-tuning both the U-Net and the text encoder or by training the U-Net from-scratch with a frozen in-domain text encoder, also achieve high scores that are only around 2 to 5\% lower than the best model, trained on multiple views. Nevertheless, we suspect that training longer would give a bigger advantage to the model using all views, which would be capable of leveraging a larger and more diverse set of training images and prompts. 
 
\subsection{Zero-shot Image-Image Retrieval}

This evaluation is similar to the conventional content-based image retrieval setting in which we search for images of a particular category using a representative query image. For evaluation, a group of query images and a larger collection of candidate images, each with a class label, are given to a pretrained CNN encoder. Each query and candidate image is encoded with this encoder, and then for each query, we rank all candidates by their cosine similarities to the query in descending order. A point of accuracy is given if a retrieved candidate image has the same category (same abnormality label) than the query image. \\

As pretrained visual encoder, we use conVIRT~\cite{zhang2020contrastive}, a multimodal model trained using contrastive learning methods using radiology reports and chest x-rays. For the images, we use the MIMIC-CXR test-set. We filter this pool of images by only keeping images whose report only contains one abnormality that is either atelectasis, consolidation, cardiomegaly, edema, fracture, pleural effusion, pneumonia or pneumothorax (we discard reports with multiple labels) . At the end of the filtering, we keep 200 query images and 400 candidate images. We focus our evaluation on retrieval precision, and evaluate our models with Precision$@k$ metrics where $k = 5, 10, 50$. Note that for our evaluations, the query images are the images generated by our different models (given the selected reports as described).\\

Compared to the previously described RRG task, the image-image retrieval task allows us to better discriminate our different trained variants, seeing the scores reported in \cref{score-tabular-fact-orient}. The model trained on multiple views outperforms the second best variant that generates PA images only, both further fine-tuning the text encoder and the U-Net for 60k steps, on all precision scores by a 7 to 13\% improvement. Though the models trained with domain-specific text encoders are not the best performing ones, we account for the fact that they only fine-tune the U-Net and keep the text encoder frozen. We can notice that they outperform the variant that uses the same fine-tuning approach but with the base CLIP text encoder, further suggesting that correctly fine-tuning these domain-specific text encoders, instead of keeping them frozen, is a promising approach to improve the generation capabilities of our models.

\subsection{Zero-shot Image-Text Retrieval}

This task is similar to the Image-Image scenario, with the difference that a query image embedding is map into a textual embedding space to retrieve the most likely impression given the image. For this experiments, we chose the CXR-RePaiR model~\cite{endo2021retrieval}: a retrieval-based radiology report generation approach using a pre-trained contrastive language-image model. This model is trained using a CLIP-style approach: the objectives maximizes the similarity between the ground-truth text and image pairs embeddings. We also use the MIMIC-CXR test-set containing paired impressions-images.\\

We still focus our evaluation on retrieval precision using Precision$@k$ with $k = 5, 10, 50$.  A point of accuracy is given if a retrieved candidate impression has the same category (same abnormality label) than the query image. In this settings, the NLG metrics are not suitable. Nonetheless, we can use the factual correctness metrics presented in Section~\ref{sec:fact_orient_rrg}. To do so, we evaluate the top-1 retrieved impression against the ground-truth impression of the image.\\

Similar to the RRG and image-image retrieval metrics, the image-text retrieval scores underline the superiority of training on data from multiple views, achieving the best performance for every metric of this task on top of the other tasks. As a new method to benchmark language models, training along a frozen SapBERT \cite{liu-etal-2021-self} leads to improved performance compared to RadBERT \cite{chambon_cook_langlotz_2022}, suggesting that SapBERT has more and better structured in-domain knowledge about radiology reports. Though for the image-only image-image retrieval task, further fine-tuning the original SD weights systematically outperformed the SapBERT-based model, we notice that for both tasks that directly rely on text, SapBERT-based training can outperform a model that uses CLIP text encoder, even in the setting where the CLIP text encoder is fine-tuned (while SapBERT is systematically frozen). CLIP training has a strong focus on learning features that are relevant to both text and images, possibly leading to better signal for an image-image retrieval task. We hypothesize that an in-domain text encoder could better capture some fine-grained textual details when encoding the impression prompts, improving performance on tasks where text is then directly used to attribute a score.

\subsection{Qualitative evaluation}

\begin{figure}
  \centering
   \includegraphics[width=\linewidth]{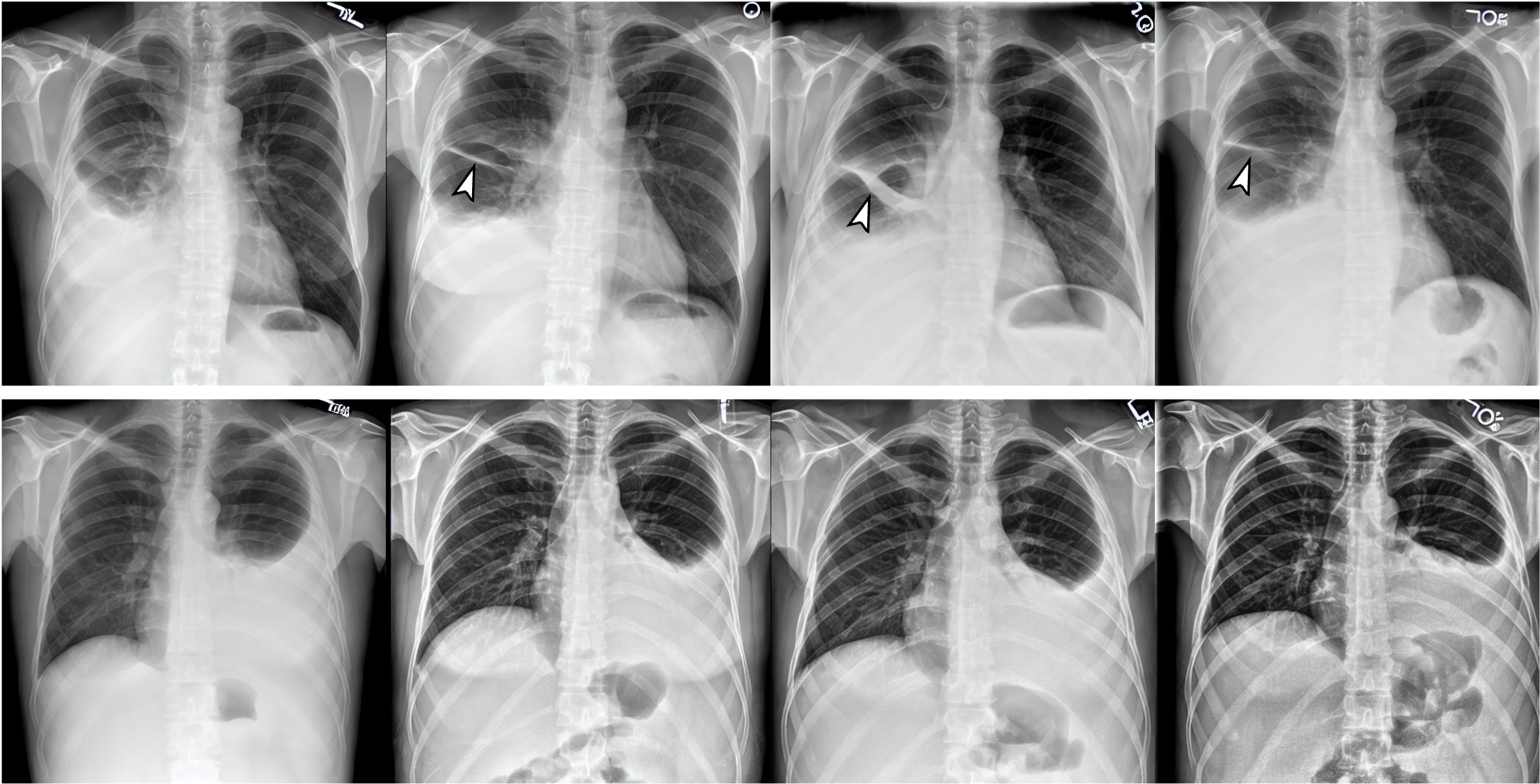}
  \caption{Intra-prompt synthetic image diversity. Four generated samples using the prompts 'Big right-sided pleural effusion with adjacent atelectasis' (top row) and 'Big left-sided pleural effusion with adjacent atelectasis' (bottom row). Note the diverse apperance of the right-sided pleural effusion with atelectasis and varying amounts of interlobar fluid (top row, white arrowheads), and the differences in contrast, with higher contrasts especially in the three left images in the bottom row, similar to real CXR when using different X-ray tube voltage settings.}
  \label{fig:diversity_cxr}
\end{figure}

In order to assess the visual quality of the generated images as well as the alignment of the radiological concepts with the text prompts, two radiologists with 7 and 9 years of experience in reading CXR were asked to review and rate:
\begin{itemize}
    \item 104 blinded pairs of true images and synthetic images (generated from the corresponding prompts, from a balanced sample of p19 test set), rating them on a scale from -2 (synthetic image is more realistic) to 2 (true image is much more realistic).
    \item 107 pairs of synthetic images and original prompts, rating them on a scale from -2 (the synthetic image does not correspond at all to the conditioning prompt), -1 (the image does not show all of the findings mentioned in the impression section), 0 (the image shows the most salient finding of the prompt but not all aspects), 1 (the image shows most of the aspects mentioned in the prompt) to 2 (the synthetic image aligns perfectly with the prompt). 
\end{itemize}

In terms of diversity, and compared to previous work fine-tuning the SD pipeline in a few-shot setting\cite{chambon2022adapting}, the generated outputs feature a broad variability of features per prompt (\cref{fig:diversity_cxr}).

For the first experimental setup, the average ratings given by the two radiologists were on average $1.67\pm 0.63$ and $1.81\pm0.46$. While the general image appearance was visually very similar to real CXR, some aspects like electrodes or other device components almost always contained unrealistic features (e.g. discontinuations or streaks), which made this distinction simple. In the majority of cases radiologists were thus able to confidently guess which image was true and which was synthetic.

The second experiment yielded average ratings of $0.41\pm 1.41$ and $0.29\pm1.36$. This underlines the ability of the model, on average, to fulfill the conditional prompt, although in most cases not all contents of the impression section translated correctly to the imaging domain. 

\section{Data augmentation}

To investigate the added value of creating synthetic CXR, a DenseNet-121 classifier was trained from scratch on varying splits of real training data (R), synthetic data (S) generated from a model trained on the p10 subset of MIMIC-CXR, and synthetic data generated on data from p10-p18. The task was a multi-label classification of six findings (cardiomegaly, edema, pleural effusion, pneumonia, pneumothorax and 'no finding'). The fine-tuned SD models were trained for 12.5k (p10) and 60k (p10-p18) training steps and sampled using impression sections from p10 or p10-p18. All classification models were trained using an AdamW optimizer (learning rate, 1e-3; weight decay, 1e-5), a cyclic learning rate scheduler and early stopping when the validation set AUROC did not improve for 15 epochs. The models with the highest validation AUROC were used to classify the p19 test set. \\

Training exclusively on 1.1k synthetic images derived from a model fine-tuned on the small dataset, a drop of 0.04 in AUROC was observed compared to the baseline. Training exclusively on 5$\times$ the initial amount of synthetic data yielded a small improvement over the baseline (AUROC +0.02). Augmenting real data from the small dataset with the same amount of synthetic data (1.1k) led to a moderate improvement (AUROC +0.04), but counter-intuitively, further augmenting the small dataset with 5.5k synthetic samples led to a smaller increase. Adding more training data (30k) improved the classification performance (AUROC +0.09), as did training exclusively on 30k synthetic samples trained on the larger dataset (AUROC +0.07). Finally, the highest improvement in classification performance was reached by training on a combination of real and synthetic data (AUROC +0.11 vs. AUROC 0.73 in the baseline setup). See \cref{tab:data-augmentation} for details. 

\begin{table}[]
\begin{tabular}{@{}llcccc@{}}
\toprule
Experiment    & \multicolumn{2}{c}{Training Data} & AUROC       & Accuracy \\ \cmidrule(lr){2-3}
              & Real            & Synth. & \multicolumn{2}{c}{(n = 5,000)}          \\ \midrule
\multicolumn{2}{l}{\textbf{Small dataset}} & & &  \\
~~ R (Baseline)             & 1.1k             &                 & 0.73        & 0.71     \\
~~ 1.1k S             &              & 1.1k                 &       0.69 ($\downarrow$0.04)  & 0.66     \\
~~ 1.1k R/S             & 1.1k             & 1.1k                 &  0.77 ($\uparrow$0.04)        &  0.76    \\
~~ 5.5k S             &                 & 5.5k               & 0.75 ($\uparrow$0.02) & 0.78     \\
~~ R + 5.5k S      & 1.1k             & 5.5k               & 0.76 ($\uparrow$0.03) & 0.77     \\
\multicolumn{2}{l}{\textbf{Big dataset}} & & &  \\
~~ 30k R         & 30k              &                 & 0.82 ($\uparrow$0.09) & 0.75     \\
~~ 30k S         &                  & 30k              & 0.80 ($\uparrow$0.07) & 0.74     \\
~~ 30k R/S & 30k              & 30k              & \textbf{0.84 ($\uparrow$0.11)} & \textbf{0.79}     \\ \bottomrule
\end{tabular}
\label{tab:data-augmentation}
\caption{Classification performance of a DenseNet-121 trained from scratch on varying splits of real and synthetic training data. 1.1k and 5.5k synthetic data was sampled from a model trained on p10 (small datasets), conditioned on prompts from p10. 30k synthetic data was sampled from a model trained on p10-p18 (big dataset), conditioned on prompts from p10-18. R: real data, S: synthetic data, AUROC: area under the ROC curve, calculated on a 5k sample of the p19 test set.}
\end{table}

\section{Distilling in-domain knowledge and potential catastrophic forgetting}

\begin{figure*}
  \centering
   \includegraphics[width=\linewidth]{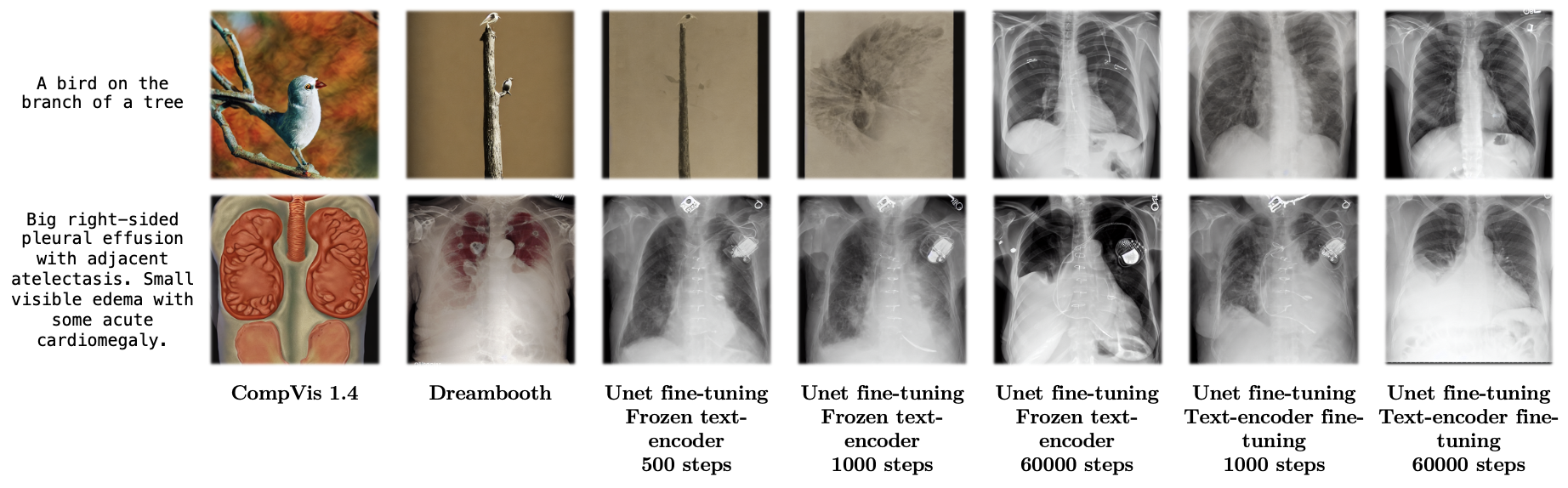}
  \caption{Comparing baselines and various models, fine-tuning either the U-Net alone or both the U-Net and the text encoder, based on a general-domain prompt and an in-domain one.}
  \label{fig:catastrophicforgetting}
\end{figure*}

By fine-tuning the U-Net alone or both the U-Net and the text encoder on the Chest X-ray domain, we can distil in-domain knowledge into the components of the SD model. The quantitative and qualitative analysis of the previous sections demonstrate the capabilities of the model at learning visual representations that achieves high fidelity, diversity and medical correctness. Nevertheless, this does not measure the performance of the model on the general domain it was initially trained on, as well as the the performance of the isolated components of the fine-tuned SD model. Looking at \cref{fig:catastrophicforgetting}, we notice that the current setting leads to a catastrophic forgetting of the knowledge previously acquired. In particular, fine-tuning both the text encoder and the U-Net accelerates the learning of in-domain concepts but also the forgetting of the previous domain knowledge.\\

As in our current setting we do not leverage any specific method to control catastrophic forgetting \cite{DBLP:journals/corr/abs-1801-06146}, we observe that as the model is introduced to new unseen images, its weights get rapidly altered leading to knowledge collapse. \cref{tab:textencoder} underlines the drop of performance for the $CheXpert@10$ score \cite{chambon2022adapting}, that measure the in-domain knowledge of the text encoder, for the first 1000 steps, where the macro-averaged score drops between 20\% and 40\%. 

Finally, looking at \cref{tab:textencoder}, we notice that the SD task can improve the performance of the text encoder on an in-domain task, in this case as measured by the macro-averaged $CheXpert@10$ score. Using learning rate $\text{5e-5}$, fine-tuning the model for 1k and 12.5k lead to a drop of performance of 43\% and 55\%, that gets partially recovered at 60k to only 25\%. In the specific case of pneumothorax, the knowledge of the model is even totally recovered or improved, if using a stronger learning rate of 1e-4. Though the strong catastrophic forgetting of our current training setting mitigates the amount of added knowledge of the text encoder at the end of the SD fine-tuning, this measure shows that such improvement of representation capabilities already happen for some abnormalities. Coupled with methods that reduce the catastrophic forgetting phenomenon, SD fine-tuning could be used to further adapt text-encoders for a particular domain.

\begin{table}
  \centering
  \begin{tabular}{@{}lcccc@{}}
    \toprule
    Model & Frac. & Ple.oth. & Pneux & Macro \\
    \midrule
    \textbf{Baseline} & & & &\\
    ~ original SD & 61.6 & 2.9 & 48.6& 40.5\\
    \textbf{1k training steps} & & & &\\
    ~ lr 5e-5 & 55.4 & 2.1 & 24& 23.2\\
    ~ lr 1e-4 & 58 & 2.1 & 55.4& 31.3\\
    \textbf{12.5k training steps} & & & &\\
    ~ lr 5e-5 & 37.4 & 2.9 & 17.2& 18.1\\
    ~ lr 1e-4 & 43.4 & 2.9 & 49.8& 28.4\\
    \textbf{60k training steps} & & & &\\
    ~ lr 5e-5 & 38.2 & 5 & 48.4& 30.2\\
    ~ lr 1e-4 & 50 & 6.4 & 59.8& 34.4\\
    ~ lr 5e-5, Rnd U-Net init & 15.4 & 0 & 11& 16.1\\
    ~ lr 5e-5, PA/AP/LAT & 29.6 & 6.4 & 46.8& 27.1\\
    \bottomrule
  \end{tabular}
  \caption{Class-wise (Fracture, Pleural other, Pneumothorax) and total macro-averaged $CheXpert@10$ scores \cite{chambon2022adapting} computed for the text encoder of various fine-tuned SD models, using p19 test set. Higher scores denote better capabilities of the model at clustering reports per abnormality, in the latent space. All models except the \textit{PA/AP/LAT} one were trained on the PA training set. All models except \textit{Rnd U-Net init}, which has a random U-Net initialization, were further fine-tuned from the CompVis 1.4 baseline with unfrozen U-Net and text encoder.}
  \label{tab:textencoder}
\end{table}

\section{Limitations}

\methodname{} is capable of generating synthetic Chest X-ray images that can be conditioned on prompts using medical language, however, limitations of the proposed approach remain:

\begin{enumerate}
    \item The CXR images generated by \methodname{} are images and not actual radiographs, and come with a limited range of gray-scale values, preventing the use of operations like realistic windowing. They should not be regarded as a replacement of actual chest x-ray studies.
    \item Only one dataset (MIMIC-CXR), from a single institution, was used to fine-tune and evaluate \methodname{}. Measuring the cross-institutional performance and robustness of the model, as well as further training it on multi-institutional data, could make the model more generalizable, improve its understanding of conditioning prompts and increase the diversity of the synthetic images.
    \item Only the impression sections from the radiology reports associated with each image were used to train the model. These sections might not capture every aspect of the image, and training the model on other sections (especially findings) could enable more fine-grained conditioning capabilities.
    \item We noticed that the model was prone to overfitting when trained on small datasets of a few hundreds of images. Follow-ups that study fine-tuning under limited data constraints could benefit from detailed measures of overfitting.
    \item Fine-tuning both the U-Net and the text encoder leads to catastrophic forgetting phenomenons. Developing methods that can limit these from happening could help speed up the training process as well as retain knowledge previously acquired.

\end{enumerate}

\section{Conclusion and future work}

The latent diffusion model Stable Diffusion, pretrained on billions of natural image-text pairs, can be domain-adapted to generate high-fidelity yet diverse medical CXR images. As established in this work, by exploring various fine-tuning approaches and evaluation settings, the best-performing model (\methodname{}, a portmanteau of "Roentgen", as a hommage to one of the pioneers of radiology, Wilhelm Conrad Röntgen, and "Generator") allows fine-grained control over the generated output by using free-form, natural language text as input, including relevant medical vocabulary. 

The domain-adapted models can display a large variety of radiological findings (e.g., pleural effusions, atelectases, pneumothoraces) beyond the scope of the usual class-conditioned setting of previous generative approaches. Additionally, the image appearance of these findings can be modified using natural (e.g., by prompting for a specific side or variations in size) and medical language (e.g. requestion findings like "consolidation" or "pleural effusion"), without being explicitly trained.

The best performance was observed after jointly fine-tuning both pretrained U-Net and the text encoder. Replacing the frozen CLIP text encoder with a domain-specific text encoder improves performance when training the U-Net from-scratch, underlining potential training speed-up when aiming for domain adaptation.
 
We developed an evaluation framework that can assess medical correctness of synthetic images with various downstream applications, such as radiology report generation or image-image and image-text retrieval, allowing us to compare different generative model in terms of clinical consistency and completeness. In addition, we used fine-tuned stable diffusion for downstream applications to either data augment training datasets or replace real training data with purely synthetic data, leading to improvements in both cases and demonstrating the added-value of such models. Finally, we measured the ability of stable diffusion fine-tuning to distill in-domain knowledge into its components, in particular the text-encoder, improving its representation capabilities on in-domain data. 
 
Building upon these findings, future research will focus on expanding the work to other study types and modalities, furthering the medical information a fine-tuned stable diffusion model could retain. In particular, we would like to further investigate fine-tuning strategies that would allow to limit catastrophic forgetting and therefore benefit from previously acquired knowledge, through the use of domain-specific text-encoders for instance.

\section*{Acknowledgements}

Research reported in this publication was made possible in part by the \textit{National Institute of Biomedical Imaging and Bioengineering (NIBIB)} of the \textit{National Institutes of Health}, which funded PC under contracts
75N92020C00008 and 75N92020C00021. 
CB received support from the Swiss Society of Radiology and the Kurt and Senta Herrmann-Foundation, independent of this work. We acknowledge support by Stability.AI in providing computational support for this work. We acknowledge the support of this work by the Wu Tsai Human Performance Alliance at Stanford University and the Joe and Clara Tsai Foundation. 
We would also like to acknowledge the help of Arjun Desai in proofreading this manuscript.

{\small
\bibliographystyle{ieee_fullname}
\bibliography{egbib}
}

\appendix
\section{Details of the factually-oriented metrics}\label{app:detail_fact_oriented}

\textbf{$\text{fact}_\text{ENT}$}~\cite{miura2021improving} \quad A named entity recognizer (NER) is applied to the generated report $\hat{y}$ and the corresponding reference $y$, giving respectively two sets of extracted entities $\mathbb{E}_{\hat{y}}$ and $\mathbb{E}_{{y}}$. $\text{fact}_\text{ENT}$ is defined as the harmonic mean of precision and recall between the two sets $\mathbb{E}_{\hat{y}}$ and $\mathbb{E}_{{y}}$. The clinical model of Stanza~\cite{qi2020stanza} is used as NER. \\

\textbf{$\text{fact}_\text{ENTNLI}$}~\cite{miura2021improving} \quad  This score is an extension of \textbf{$\text{fact}_\text{ENT}$} with Natural Language Inference (NLI). Here, an entity $e$ of $\mathbb{E}_{\hat{y}}$ is not automatically considered correct if present in $\mathbb{E}_{{y}}$. To be considered valid, the sentence $s_{\hat{y}}$ containing entity $e$ must not present a contradiction with its counterpart sentence $s_{{y}}$ in the reference report. The counterpart sentence $s_{{y}}$ in the reference report is the sentence with the highest BERTScore~\cite{zhang2019bertscore} against $s_{\hat{y}}$. The NLI model outputs whether sentence $s_{{y}}$ is a contradiction of $s_{\hat{y}}$. We use the NLI model weights of \cite{miura2021improving}, which relies on a BERT-architecture. \\

\textbf{F$_1$CheXbert}~\cite{zhang2020optimizing} \quad This score uses CheXbert~\cite{smit2020combining}, a Transformer-based model trained to output abnormalities (fourteen classes) of chest X-rays given a radiology report as input. F$_1$CheXbert is the F1-score between the prediction of CheXbert over the generated report $\hat{y}$ and the corresponding reference $y$. To be consistent with previous works, the score is calculated over 5 observations: atelectasis, cardiomegaly, consolidation, edema and pleural effusion. \\

\end{document}